%% file: acl2021.tex
\pdfoutput=1
\documentclass[11pt]{article}

\usepackage[]{acl2021}

\usepackage{times}
\usepackage{latexsym}

\usepackage[T1]{fontenc}
\usepackage[utf8]{inputenc}

\usepackage{microtype}

\usepackage{makecell}

\usepackage{microtype}

\usepackage{graphicx}
\usepackage{paralist}
\usepackage{url}
\usepackage{xspace}
\usepackage{amsfonts,amsmath}
\usepackage{booktabs}
\usepackage{setspace}
\usepackage{subcaption}
\usepackage{multirow}
\usepackage{multicol}

\usepackage{hyperref}
\usepackage{amssymb}% http://ctan.org/pkg/amssymb
\usepackage{pifont}% http://ctan.org/pkg/pifont
\usepackage{color, colortbl}
\usepackage{bbm}

\newcommand\blfootnote[1]{%
  \begingroup
  \renewcommand\thefootnote{}\footnote{#1}%
  \addtocounter{footnote}{-1}%
  \endgroup
}

\newcommand*{\scale}[2][4]{\scalebox{#1}{$#2$}}

\makeatletter
\newcommand{\thickhline}{%
    \noalign {\ifnum 0=`}\fi \hrule height 1pt
    \futurelet \reserved@a \@xhline
}
\usepackage{balance}
\makeatother

\title{\vspace*{-0.4in}
{\small \hfill Accepted to ACL 2022} \\
\vspace{0.2in}
Diversifying Content Generation for Commonsense Reasoning \\ with Mixture of Knowledge Graph Experts}
\author{{\bf Wenhao Yu$^{\clubsuit}$, Chenguang Zhu$^{\spadesuit}$, Lianhui Qin$^{\heartsuit}$,} \\ \bf{Zhihan Zhang$^{\clubsuit}$, Tong Zhao$^{\clubsuit}$, Meng Jiang$^{\clubsuit}$} \\
\normalsize{ $^\clubsuit$University of Notre Dame $^\heartsuit$University of Washington} \\ 
\normalsize{$^\spadesuit$Microsoft Cognitive Services Research} \\
{\normalsize{\tt $^\clubsuit$\{wyu1, zzhang23, tzhao2, mjiang2\}@nd.edu}} \\
\normalsize{ {\tt $^\spadesuit$chezhu@microsoft.com} \ \ $^\heartsuit$\tt lianhuiq@cs.washington.edu}
}

\begin{document}
\maketitle

\blfootnote{$\S$ Codes of our model and baselines are available at \url{https://github.com/DM2-ND/MoKGE}. }

\begin{abstract}
\input{0abstract}

\end{abstract}

\section{Introduction}
\label{sec:introduction}
\input{1introduction}

\section{Related Work}
\label{sec:related}
\input{2related}

\section{Proposed Method}
\label{sec:method}
\input{3method}

\section{Experiments}
\label{sec:Experiments}
\input{4experiments}

\section{Future Directions}
\input{5future}

\section{Conclusions}
\label{sec:conclusions}
\input{5conclusions}

\section*{Acknowledgements}
The work is supported by National Science Foundation IIS-1849816, CCF-1901059, and IIS-2119531.

\balance
\bibliography{reference}
\bibliographystyle{acl_natbib}

% \clearpage
% \appendix
% \section{Appendix}
% \input{6appendix}

\end{document}

%% file: 0abstract.tex
Generative commonsense reasoning (GCR) in natural language is to reason about the commonsense while generating coherent text. 
Recent years have seen a surge of interest in improving the generation quality of commonsense reasoning tasks. Nevertheless, these approaches have seldom investigated diversity in the GCR tasks, which aims to generate alternative explanations for a real-world situation or predict all possible outcomes. Diversifying GCR is challenging as it expects to generate multiple outputs that are not only semantically different but also grounded in commonsense knowledge.
In this paper, we propose MoKGE, a novel method that diversifies the generative reasoning by a mixture of expert (MoE) strategy on commonsense knowledge graphs (KG). A set of knowledge experts seek diverse reasoning on KG to encourage various generation outputs. Empirical experiments demonstrated that MoKGE can significantly improve the diversity while achieving on par performance on accuracy on two GCR benchmarks, based on both automatic and human evaluations.

%% file: 1introduction.tex
An important desideratum of natural language generation (NLG) is to produce outputs that are not only correct but also diverse~\cite{tevet2021evaluating}. 
The term ``diversity'' in NLG is defined as the ability of a generative model to create a set of possible outputs that are each valid given the input and vary as widely as possible in terms of \textit{content, language style, and word variability}~\cite{gupta2018deep}. This research problem is also referred as \textit{one-to-many generation}~\cite{shen2019mixture,cho2019mixture,yu2021sentence,shen2022diversified}. 

Diversity in NLG has been extensively studied for various tasks in the past few years, such as machine translation~\cite{shen2019mixture} and paraphrase generation~\cite{gupta2018deep}.
In these tasks, output spaces are constrained by input context, i.e., the contents of multiple outputs should be similar, and globally, under the same topic.
However, many NLG tasks, e.g., generative commonsense reasoning, pose unique challenges for generating multiple reasonable outputs that are \textit{semantically different}.
% from each other.
% however, to the best of our knowledge, we are the first work to investigate content diversity in the generating commonsense reasoning tasks. 

\begin{figure}[t]
    \centering
    {\includegraphics[width=0.46\textwidth]{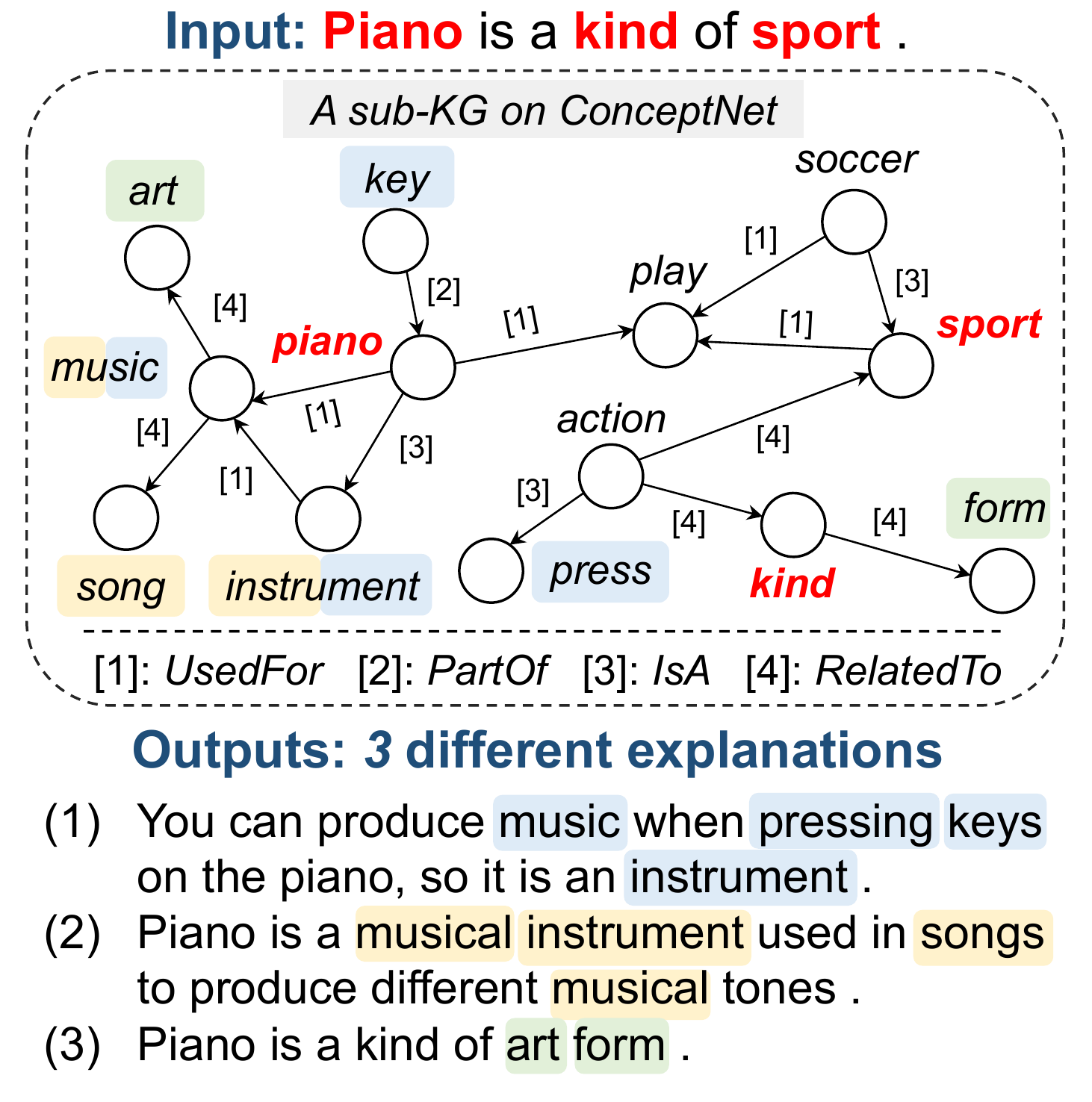}}
    \vspace{-0.15in}
    \caption{An example of diverse commonsense explanation generation. It aims at generating multiple reasonable explanations given a counterfactual statement. Relevant concepts on the commonsense KG (in shade) can help to perform diverse knowledge reasoning.}
    \label{fig:intro}
\end{figure}

Figure \ref{fig:intro} shows an example in the commonsense explanation generation (ComVE) task. The dataset has collected explanations to counterfactual statements for sense-making from three annotators~\cite{wang2020semeval}.
From the annotations, we observed that different annotators gave explanations to the unreasonable statement from different perspectives to make them diverse in terms of content, e.g., wrong effect and inappropriate usage.%, erroneous object type.

In order to create diversity, existing methods attempted to produce \textit{uncertainty} by introducing random noise into a latent variable~\cite{gupta2018deep} or sampling next token widely from the vocabulary~\cite{holtzman2020curious}.
However, these methods were not able to explicitly control varying semantics units and produce outputs of diverse content.
% Existing Seq2Seq model provides limited reasoning by simply transforming their hidden state given the input text.
Meanwhile, the input text alone contains too limited knowledge to support diverse reasoning and produce multiple reasonable outputs~\cite{yu2020survey}.
As an example, Table \ref{tab:intro} shows the human evaluation results on two GCR tasks. While human annotators were able to produce 2.60 different yet reasonable explanations on the ComVE dataset, one SoTA diversity-promoting method (i.e., nucleus sampling~\cite{holtzman2020curious}) could produce only 2.15 reasonable explanations. 

% As shown in Table \ref{tab:intro}, the performance of the state-of-the-art (SOTA) diversity-promoting methods is still much worse than human performance, based on human evaluation. For example, human annotators were able to produce 2.60 different yet reasonable explanations on average in the ComVE dataset~\cite{wang2020semeval}, while the SOTA~\cite{holtzman2020curious} could produce only 2.15. In comparison, our method exceeds the human performance and creates 2.63 different reasonable explanations on average.

To improve the diversity in outputs for GCR tasks, we investigated the ComVE task and found that 75\% of the concepts (nouns and verbs) in human annotations were among 2-hop neighbors of the concepts contained in the input sequence on the commonsense KG ConceptNet\footnote{ConceptNet: \url{https://conceptnet.io/}}.
Therefore, to produce diverse GCR, our idea is enabling NLG models to reason from different perspectives of knowledge on commonsense KG and use them to generate diverse outputs like the human annotators.

% Commonsense reasoning, the ability of making logical assumptions about ordinary scenes in human daily lives, has long been acknowledged as a critical bottleneck of artificial intelligence~\cite{lin2020commongen}.
% Generative reasoning (GR), not only requires to reasonably associate additional commonsense knowledge, but also needs machines to complete a coherent scenario through language generation technologies~\cite{liu2021kg}.

% Existing GR work explored leveraging commonsense knowledge graph via reasoning relational paths or enriching conceptual representation by graph neural networks to improve the generation quality~\cite{ji2020language,liu2021kg}. Nevertheless, an important desideratum of the GR task is to produce outputs that are not only correct, but also diverse~\cite{tevet2020evaluating}. 

\begin{table}[t]
\caption{Under human evaluation, the performance of existing diversity promoting methods is still far from that of humans. Our method MoKGE can exceed the human performance on the ComVE task.}
\vspace{-0.15in}
\begin{center}
\scale[0.9]{\begin{tabular}{lcc}
\toprule
& ComVE & $\alpha$-NLG \\
\midrule
% Avg. input words & 7.7 & 17.4 \\
% Avg. output words & 9.0 & 10.8 \\
Avg. \# human references & 3.00 & 4.20 \\
% \# instance & 10,000 & 5,339 \\
% \hline
% Jaccard similarity ($\Downarrow$) & & \\ 
% \;Human & 47.33 & 30.12  \\
% \;SOTA [ref?] & 57.77 & 44.79 \\
% \;Ours & 50.38 & 20.04 \\
% \hline
% Self-BLEU-4 ($\Downarrow$) &&  \\
% \;Human & 7.83 & 8.47 \\
% \;SOTA [ref?] & 41.86 & 22.50 \\
% \;Ours & 24.23 & 19.69 \\
\midrule
Avg. \# meanings ($\Uparrow$) &&\\
\;Human references & \underline{2.60} & \textbf{3.79} \\
\;Nucleus sampling & 2.15 & 3.35 \\
\;MoKGE (our method) & \textbf{2.63} & \underline{3.72} \\
% \midrule
% ComVE v.s. $\alpha$-NLG & Hard & Easy \\
\bottomrule
\end{tabular}}
\end{center}
% \footnote{Concept is defined as a word appears in the ConceptNet.}
\vspace{-0.1in}
\label{tab:intro}
\end{table}

%One potential solution is to exploit semantic information from the commonsense knowledge graph and to perform reasoning over different multi-hop relational paths to generate diverse outputs. 
Thus, we present a novel \underline{\textbf{M}}ixture \underline{\textbf{o}}f \underline{\textbf{K}}nowledge \underline{\textbf{G}}raph \underline{\textbf{E}}xpert (MoKGE) method for diverse generative commonsense reasoning on KG. MoKGE contains two major components: (i) a knowledge graph (KG) enhanced generative reasoning module to reasonably associate relevant concepts into the generation process, and (ii) a mixture of expert (MoE) module to produce diverse reasonable outputs.           
Specifically, the generative reasoning module performs compositional operations on KG to obtain structure-aware representations of concepts and relations. Then, each expert uses these representations to seek different yet relevant sets of concepts and sends them into a standard Transformer model to generate the corresponding output. To encourage different experts to specialize in different reasoning abilities, we employ the stochastic hard-EM algorithm by assigning full responsibility of the largest joint probability to each expert.
% To mitigate the lack of ground truth annotation about which concept node should be selected, we use the overlapping concepts between concepts appearing in the output sequence and concepts on the knowledge graph as a simple proxy for the ground-truth supervision. 

We conducted experiments on two GCR benchmarks, i.e., commonsense explanation generation and abductive commonsense reasoning. Empirical experiments demonstrated that our proposed MoKGE can outperform existing diversity-promoting generation methods in diversity, while achieving on par performance in quality.

To the best of our knowledge, this is the first work to boost diversity in NLG by diversifying knowledge reasoning on commonsense KG.

% \begin{itemize}
%     \item To the best of our knowledge, we are the first to boost diversity in NLG by diversifying the knowledge reasoning on commonsense KG.
%     \item We propose a novel method called MoKGE to exploit semantic information and perform reasoning over the commonsense KG to generate diverse outputs.
%     \item Experiments on two generative reasoning benchmarks demonstrated that our MoKGE outperformed existing diversity-promoting generation methods based on diversity evaluation. At the same time, it achieved on par performance based on quality evaluation. 
% \end{itemize}

%% file: 2related.tex
\begin{figure*}[th]
    \centering
    {\includegraphics[width=1.0\textwidth]{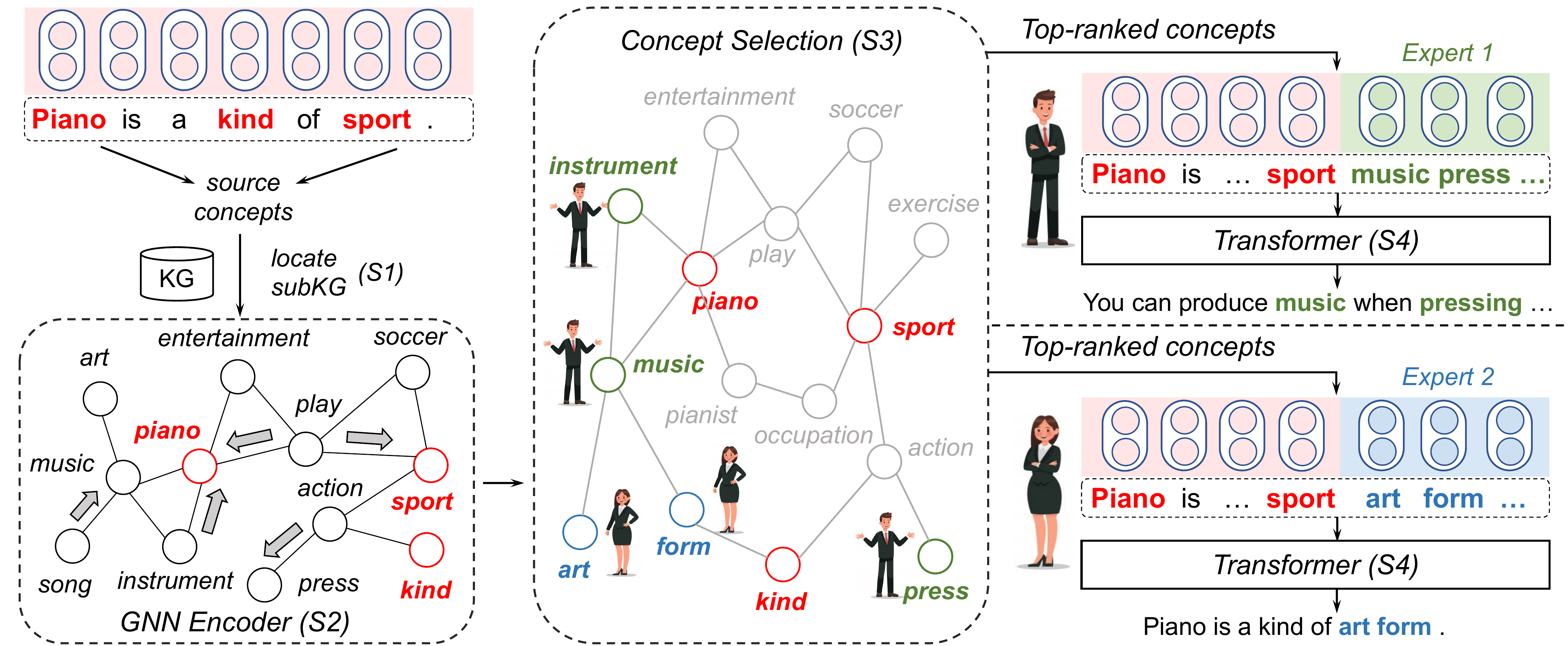}}
    \vspace{-0.2in}
    \caption{The overall architecture of MoKGE. 
    % There are 2 experts in the figure, but the number of experts can vary. In the experiments, we have tested 3, 4 and 5 experts. 
    The MoKGE consists of four steps: (S1) the model constructs a sequence-associated subgraph from the commonsense KG; (S2) a relational-GCN iteratively updates the representation of a concept node by aggregating information from its neighboring nodes and edges; (S3) each knowledge expert selects different salient concepts that should be considered during generation; (S4) the model generates the outputs by integrating the token embeddings of the input sequence and the top-ranked entities.}
    \label{fig:framework}
\end{figure*}

\subsection{Diversity Promoting Text Generation}

Generating multiple valid outputs given a source sequence has a wide range of applications, such as machine translation~\cite{shen2019mixture}, paraphrase generation~\cite{gupta2018deep}, question generation~\cite{cho2019mixture}, dialogue system~\cite{dou2021multitalk}, and story generation~\cite{yu2021sentence}. 
For example, in machine translation, there are often many plausible and semantically equivalent translations due to information asymmetry between different languages~\cite{lachaux2020target}. 
% in dialogue system, given a similar observed inputs, there may exist many valid responses~\cite{zhao2017learning}. 
% in question generation, a document can be used to ask several different question that covers different perspectives~\cite{cho2019mixture,wang2020diversify}.
% Besides, paraphrase generation aims to generate sentences with the same meaning as an input sentence but with different lexicon or syntax~\cite{gupta2018deep}.

Methods of improving diversity in NLG have been explored from various perspectives.
Sampling-based decoding is one of the most effective solutions to improve diversity. %~\cite{fan2018hierarchical,holtzman2020curious}
For example, nucleus sampling~\cite{holtzman2020curious} samples next tokens from the dynamic nucleus of tokens containing the vast majority of the probability mass, instead of decoding text by maximizing the likelihood.
Another line of work focused on introducing random noise~\cite{gupta2018deep} or changing latent variables~\cite{lachaux2020target} to produce uncertainty. %, e.g., \citet{gupta2018deep} employed a variational auto-encoder (VAE) to generate diverse paraphrases according to the input noise. 
In addition, \citet{shen2019mixture} adopted a mixture of experts to diversify machine translation, where a minimum-loss predictor is assigned to each source input. \citet{shi2018toward} employed an inverse reinforcement learning approach for unconditional diverse text generation.

However, no existing work considered performing diverse knowledge reasoning to generate multiple reasonable outputs of different contents.

\subsection{Knowledge Graph for Text Generation}

Incorporating external knowledge is essential for many NLG tasks to augment the limited textual information~\cite{yu2020survey,dong2021injecting,yu2022dict}.
Some recent work explored using graph neural networks (GNN) to reason over multi-hop relational knowledge graph (KG) paths%, represent structured summaries and highlight the proximity of relevant concepts
~\cite{zhou2018commonsense,jiang2019role,zhang2020grounded,wu2020diverse,yu2021kg,zeng2021enhancing}. 
For example, \citet{zhou2018commonsense} enriched the context representations of the input sequence with neighbouring concepts on ConceptNet using graph attention. 
% \citet{zhang2020grounded} proposed ConceptFlow that represented the potential conversation flow as traverses in the concept space along commonsense relations; 
% \citet{guan2019story} proposed incremental encoding with multi-source attention to incorporate knowledge graph for concepts in the story context.
\citet{ji2020language} performed dynamic multi-hop reasoning on multi-relational paths extracted from the external commonsense KG.
Recently, some work attempted to integrate external commonsense knowledge into generative pretrained language models~\cite{guan2020knowledge,bhagavatula2020abductive,liu2021kg}. 
For example, 
\citet{guan2020knowledge} conducted post-training on sythetic data constructed from commonsense KG by translating triplets into natural language texts using templates.
% \citet{bhagavatula2020abductive} transferred embeddings of COMeT, a GPT-2 model fine-tuned to generate the tail entity of a triple in commonsense knowledge graph, into another GPT-2 model for generation. 
\citet{yu2020survey} wrote a comprehensive survey for more detailed comparisons of different knowledge graph enhanced NLG methods.

% \begin{figure*}[th]
%     \centering
%     {\includegraphics[width=1.0\textwidth]{figures/framework.pdf}}
%     \vspace{-0.25in}
%     \caption{The overall architecture of KG-enhanced generative reasoning. A sequence-associated subgraph is firstly retrieved from the commonsense KG given the source input. Then, the graph neural network (GNN) iteratively updates the representation of a concept node by aggregating information from its neighboring nodes and edges. Next, the model selects salient concepts that should be considered during generation. Finally, the model generates the explanation by integrating the token embeddings of both the input sequence and the top-ranked entities.}
%     \label{fig:framework}
%     \vspace{-0.1in}
% \end{figure*}

%% file: 3method.tex
\noindent\textbf{Problem formulation.} In this paper, we focus on diversifying the outputs of generative commonsense reasoning (GCR) tasks, e.g. commonsense explanation generation and abductive commonsense reasoning. 
These tasks require \textit{one-to-many} generation, i.e., creating a set of reasonable outputs that vary as widely as possible in terms of contents, language style and word variability.
Formally, given a source input $x$, our goal is to model a conditional distribution for the target outputs $p(y|x)$ that assigns high values to $\{ p(y_1|x), \cdots, p(y_K|x) \}$ for $K$ mappings, i.e.,  $ \{ x \rightarrow y_1, \cdots, x \rightarrow y_K \}$. Meanwhile, the outputs $\{ y_1, \cdots, y_K \}$ are expected to be diverse with each other in terms of \textit{contents}.

\vspace{0.05in}
Existing diversity-promoting methods only varied the language styles and failed to perform different knowledge reasoning to generate diverse contents~\cite{cho2019mixture,shen2019mixture,holtzman2020curious}. Here, incorporating commonsense KG is essential for the generative reasoning (GR) tasks because the KG cannot only augment the limited information in the input text, but also provide a rich searching space for knowledge reasoning.
Therefore, we propose to employ commonsense KG to play the central role of performing diverse knowledge reasoning, then use different sets of selected concepts to produce diverse outputs.

\vspace{0.05in}
\noindent\textbf{Model Outline.}
Our model has two major components: (i) a knowledge graph (KG) enhanced generative reasoning module to reasonably associate relevant concepts and background into the generation process, and (ii) a mixture of expert (MoE) module to diversify the generation process and produce multiple reasonable outputs.

% To empower the generation model to produce multiple reasonable outputs, we employ a mixture of expert (MoE) module to model uncertainty and generate diverse outputs. While the MoE models have primarily been explored as a means of increasing model capacity, they are also considered as a natural way of boosting diverse generation process~\cite{shen2019mixture,cho2019mixture}.
% Formally, the MoE module introduces a multinomial latent variable $z \in \{1, \cdots, K\}$, and decomposes the marginal
% likelihood as the following format: 
% \begin{equation}
%     p(y|x, \mathcal{G}_x) = \sum^K_{z=1} p(z|x, \mathcal{G}_x)p(y|z, x, \mathcal{G}_x) , \nonumber
% \end{equation}

\subsection{KG-enhanced Generative Reasoning}

The KG-enhanced generative reasoning module is illustrated in Figure \ref{fig:framework}. It consists of four steps.
First, a sequence-associated subgraph is retrieved from the KG given the input sequence ($\S$\ref{sec:sub-construct}). Then, a multi-relational graph encoder iteratively updates the representation of each node by aggregating information from its neighboring nodes and edges ($\S$\ref{sec:graph-encoder}). Next, the model selects salient concepts that should be considered during generation ($\S$\ref{sec:concept-selection}). Finally, the model generates outputs by integrating the token embeddings of both the input sequence and the top-ranked concepts ($\S$\ref{sec:generator}).

\subsubsection{Sequence-aware subgraph construction}
\label{sec:sub-construct}

To facilitate the reasoning process, we resort to an external commonsense knowledge graph $\mathcal{G} = \{ \mathcal{V}, \mathcal{E} \}$, where $\mathcal{V}$ denotes the concept set and $\mathcal{E}$ denotes the edges with relations. Since direct reasoning on the entire graph is intractable, we extract a sequence-associated subgraph $\mathcal{G}_{x} = \{ \mathcal{V}_{x}, \mathcal{E}_{x} \}$, where $\mathcal{V}_{x}$ consists of the concepts extracted from the input sequence (denoted as $C_{x}$) and their inter-connected concepts within two hops, i.e., $\mathcal{V}_{x} = \{C_x \cup \mathcal{N}(C_x) \cup \mathcal{N}(\mathcal{N}(C_x)) \}$. For example, in Figure \ref{fig:framework}, $C_x = \{\text{piano}, \text{sport}, \text{kind} \}$ and $\mathcal{V}_{x} = \{\text{piano}, \text{sport}, \text{kind}, \text{art}, \text{music}, \text{press}, ... \}$.
% We only consider concepts with 1-gram surface texts.
Next, the generation task is to maximize the conditional probability
$p(y|x, \mathcal{G}_{x})$.

\subsubsection{Multi-relational graph encoding}
\label{sec:graph-encoder}

%As recent advances of graph neural networks (GNNs)~\cite{wu2020comprehensive} and Graph-to-Sequence~\cite{beck2018graph} potentiate to bridge the gap between graph representation learning and language generation, GNNs serve as an important role of integrating rich semantic and structural knowledge into text generation.
% Graph neural network (GNN) frameworks, such as graph convolution network (GCN)~\cite{kipf2017semi} and graph attention network (GAT)~\cite{velivckovic2018graph}, have been shown effective at encoding graph-structured data by aggregating node information from local neighbours. 
To model the relational information in the commonsen KG, we employ the relational graph convolutional network (R-GCN)~\cite{schlichtkrull2018modeling} which generalizes GCN with relation specific weight matrices. We follow \citet{vashishth2020composition} and \citet{ji2020language} to use a non-parametric compositional operation $\phi(\cdot)$ to combine the concept node embedding and the relation embedding. Specifically, given the input subgraph $\mathcal{G}_{x} = \{ \mathcal{V}_{x}, \mathcal{E}_{x} \}$ and an R-GCN with $L$ layers, we update the embedding of each node $v \in \mathcal{V}_{x}$ at the $(l+1)$-th layer by aggregating information from the embeddings of its neighbours in $\mathcal{N}(v)$ at the $l$-th layer:
% which consist of pairs of node $u$ and the connected relation $r$.
\begin{equation}
    \textbf{o}_v^l = \frac{1}{|\mathcal{N}(v)|}\sum_{(u, v, r) \in \mathcal{E}} \textbf{W}^{l}_{N} \phi(\textbf{h}^{l}_{u}, \textbf{h}^{l}_{r}),
\end{equation}
\vspace{-0.1in}
\begin{equation}
    \textbf{h}^{l+1}_{v} = \mathrm{ReLU}(\textbf{o}^{l}_{v} + \textbf{W}^{l}_{S}\textbf{h}^{l}_{v}),
\end{equation}
where $\textbf{h}_v$ and $\textbf{h}_{r}$ are node embedding and relation embedding.
% In addition, $\textbf{W}^l_N$ and $\textbf{W}^l_S$ are two learnable weight matrices specific to the $l$-th layer. 
We define the compositional operation as $\phi(\textbf{h}_u, \textbf{h}_r) = \textbf{h}_u - \textbf{h}_r$ inspired by the TransE~\cite{bordes2013translating}.
The relation embedding is also updated via another linear transformation: 
% parameterized by $\textbf{W}^l_R$:
\begin{equation}
    \textbf{h}^{l+1}_r = \textbf{W}^l_R \textbf{h}^l_r. 
\end{equation}
Finally, we obtain concept embedding $\textbf{h}^{L}_v$ that encodes the sequence-associated subgraph context.

\subsubsection{Concept selection on knowledge graph}
\label{sec:concept-selection}
Not all concepts in $\mathcal{G}$ appear in the outputs. Thus, we design a concept selection module to choose salient concepts that should be considered during generation.
For each concept $v \in \mathcal{V}_{x}$, we calculate
its probability of being selected by taking a multi-layer perception (MLP) on the top of graph encoder: $p_v=Pr[v \text{ is selected}|x] = \mathrm{MLP}(\textbf{h}^{L}_v)$.

% To mitigate the lack of ground truth annotation about which concept should be selected, 
To supervise the concept selection process, we use the overlapping concepts between concepts appearing in the output sequence $C_y$ and concepts in input sequence associated subgraph $\mathcal{G}_x$, i.e., $\mathcal{V}_x \cap C_y$, as a simple proxy for the ground-truth supervision. So, the concept selection loss (here only for one expert, see MoE loss in Eq.(\ref{eq:moe})) is:
\begin{align}
  \mathcal{L}_{\mathrm{concept}} = & - \Big{(} \sum_{v \in \mathcal{V}_x \cap C_y} v \log p_v \\
  & + \sum_{v \in \mathcal{V}_x - C_y} (1-v) \log (1- p_v) \Big{)}. \nonumber
%   + \sum_{v \notin C_X \cap C_Y} \log[1 - p(Z|X)] )
\end{align}
Finally, the top-$N$ ranked concepts on the subgraph $G_x$ (denoted as $v_1,...,v_N$) are selected as the additional input to the generation process.

\subsubsection{Concept-aware sequence generation}
\label{sec:generator}

We utilize a standard Transformer ~\cite{vaswani2017attention} as our generation model.
It takes the concatenation of the sequence $x$ and all the selected concepts $v_1,...,v_N$ as input and auto-regressively generates the outputs $y$. We adopt the cross-entropy loss, which can be written as:
\begin{align}
\mathcal{L}_{\text{generation}} & = - \log p(y|x, v_1, \cdots, v_N) \\ &
= -\sum_{t=1}^{|y|} \log p(y_t|x, v_1, \cdots , v_N, y_{<t}). \nonumber
\end{align}
Note that since the selected concepts do not have a rigorous order, we only apply positional encodings (used in Transformer) to the input sequence $x$.

\subsubsection{Overall objective}
We jointly optimizes the following loss:
\begin{align}
    \mathcal{L} = \mathcal{L}_{\text{generation}} + \lambda \cdot \mathcal{L}_{\text{concept}}.
    \label{eq:loss}
\end{align}
where $\lambda$ is a hyperparameter to control the importance of different tasks\footnote{We performed a hyperparameter search and found when $\lambda$ was around $0.3$, the model performed the best. Therefore, we set $\lambda=0.3$ in the following experiments.}.

\subsection{MoE-Promoted Diverse Generation}

To empower the generation model to produce multiple reasonable outputs, we employ a mixture of expert (MoE) module to model uncertainty and generate diverse outputs. While the MoE models have primarily been explored as a means of increasing model capacity, they are also being used to boost diverse generation process~\cite{shen2019mixture,cho2019mixture}.
Formally, the MoE module introduces a multinomial latent variable $z \in \{1, \cdots, K\}$, and decomposes the marginal
likelihood as follows: 
\begin{equation}
    p(y|x, \mathcal{G}_x) = \sum^K_{z=1} p(z|x, \mathcal{G}_x)p(y|z, x, \mathcal{G}_x).
\end{equation}

\noindent\textbf{Training.} We minimize the loss function (in Eq.(\ref{eq:loss})) using the MoE decomposition,
%log likelihood (i.e., the objective of concept selection and output generation) by computing the following gradient:
\begin{align}
    \nabla & \log p(y|x, \mathcal{G}_x) \label{eq:moe} \\ & = \sum_{z=1}^{K} p(z|x, y, \mathcal{G}_x)  \nonumber \cdot \nabla \log p(y, z|x, \mathcal{G}_x), 
\end{align}
and train the model with the EM algorithm~\cite{dempster1977maximum}. % by iteratively applying the E/M-step.
Ideally, we would like different experts to specialize in different reasoning abilities so that they can generate diverse outputs. 
The specialization of experts means that given the input, only one element in $\{p(y, z|x, \mathcal{G}_x)\}_{z=1}^K$ should dominate in value~\cite{shen2019mixture}. To encourage this, we employ a hard mixture model to maximize $\max_z p(y, z|x, \mathcal{G}_x)$ by assigning full responsibility to the expert with the largest joint probability. Training proceeds via hard-EM can be written as:
\begin{compactitem}
    \item E-step: estimate the responsibilities of each expert $r_{z} \leftarrow \mathbbm{1}[z=\arg\max_z p(y, z|x, \mathcal{G}_x)]$ using the current parameters $\theta$;
    \item M-step: update the parameters with gradients of the chosen expert ($r_z = \mathbbm{1}$) from E-step.
\end{compactitem}

\vspace{0.05in}
\noindent\textbf{Expert parameterization.}
Independently parameterizing each expert may exacerbate overfitting since the number of parameters increases linearly with the number of experts~\cite{shen2019mixture}. 
We follow the parameter sharing schema in \citet{cho2019mixture,shen2019mixture} to avoid this issue. This only requires a negligible increase in parameters over the baseline model that does not uses MoE. In our experiments, we compared adding a unique expert embedding to each input token with adding an expert prefix token before the input text sequence, where they achieved very similar performance.

\vspace{0.05in}
\noindent\textbf{Producing K outputs during inference.} 
In order to generate $K$ different outputs on test set, we follow~\citet{shen2019mixture} to enumerate all latent variables $z$ and then greedily decoding each token by $\hat{y}_t = \arg\max p(y|\hat{y}_{1:t-1}, z, x)$. 
In other words, we ask each expert to seek different sets of concepts on the knowledge graph, and use the selected concepts to generate $K$ different outputs.
Notably, this decoding procedure is efficient and easily parallelizable.
Furthermore, to make fair comparisons with sampling-based methods, we use greedy decoding without any sampling strategy.

% \subsubsection{Parameterization}
% An important design decision with mixture models is the degree of parameter sharing between experts. Using independently parameterized experts provides them with
% additional capacity to differentiate from one another, but may exacerbate overfitting since the number of parameters increases linearly with the number of experts. On the other hand, sharing parameters among experts may help mitigate degeneracy D1, whereby low quality experts are neglected and eventually “die” during training, since by sharing parameters even unpopular experts receive some gradients. We test different model variants using both independent and shared parameters. With independent parameterization, each expert has a different decoder network. With shared parameterization, experts use the same decoder network but the beginning-of-sentence token at the start of the target sequence is replaced with an embedded representation of the latent variable. This requires a negligible increase in parameters over the baseline model.

% \begin{equation}
%     p(Y|X) = \mathbb{E}_Z \sim p(Z|X)[p(Y|Z, X)],
% \end{equation}
% where $p(Z|X)$ is the entity selection stage and $p\theta(Y|X, Z)$ is the sequence generation stage. The factorization separates entity selection from generation so that modeling multimodality (diverse outputs) can be solely handled in the select stage and generate stage can solely concentrate on the generation task itself. 

%% file: 4experiments.tex
\subsection{Tasks and Datasets}

\noindent\textbf{Commonsense explanation generation.} It aims to generate an explanation given a counterfactual statement for sense-making~\cite{wang2019does}. We use the benchmark dataset ComVE from SemEval-2020 Task 4~\cite{wang2020semeval}. The dataset contains 10,000 / 997 / 1,000 examples for training / development / test sets, respectively. The average input/output length is 7.7 / 9.0 words.
All examples in the dataset have 3 references.

\vspace{0.02in}
\noindent\textbf{Abductive commonsense reasoning.} It is also referred as $\alpha$-NLG.
It is the task of generating a valid hypothesis about the likely explanations to partially observable past and future. We use the $\mathcal{ART}$ benchmark dataset~\cite{bhagavatula2020abductive} that consists of 50,481 / 1,779 / 3,560 examples for training / development / test sets. The average input/output length is 17.4 / 10.8 words.
Each example in the $\mathcal{ART}$ dataset has 1 to 5 references.

\subsection{Baseline Methods}

We note that as we targeted at the \textit{one-to-many} generation problem, we excluded those baseline methods mentioned in the related work that cannot produce multiple outputs, e.g., \citet{zhang2020grounded,ji2020language,liu2021kg}. 
% We only list their quality performance on two datasets in Appendix \ref{sec:other-base}. 
% We also exclude \citet{wu2020diverse,li2020knowledge} that require personalized input conditions to produce different outputs. 
Different from aforementioned methods, our MoKGE can seek diverse reasoning on KG to encourage various generation outputs \textit{without any additional conditions}.

To the best of our knowledge, we are the first work to explore diverse knowledge reasoning on commonsense KG to generate multiple diverse output sequences. Therefore, we only compared our MoKGE with existing diversity-promoting baselines without using knowledge graph. 

\vspace{0.02in}
\noindent\textbf{VAE-based method.} The variational auto-encoder (VAE)~\cite{kingma2014auto} is a deep generative latent variable model. VAE-based methods produce diverse outputs by sampling different latent variables from an approximate posterior distribution. 
CVAE-SVG (SVG is short for sentence variant generation)~\cite{gupta2018deep} is a conditional VAE model that can produce multiple outputs based an original sentence as input. 
% However, VAE-based method often suffer from a posterior collapse where the sampled latent variables are ignored~\cite{bowman2016generating}. 

\vspace{0.02in}
\noindent\textbf{MoE-based method.} Mixture models provide an alternative approach to generate diverse outputs by sampling different mixture components. We compare against two mixture of experts (MoE) implementations by \citet{shen2019mixture} and \citet{cho2019mixture}. We refer them as MoE-prompt ~\cite{shen2019mixture} and MoE-embed~\cite{cho2019mixture}.
% Besides, SELECTOR~\cite{cho2019mixture} explicitly separates diversification from generation using a general plug-and-play module. It first sampled different binary masks on the source sequence and generate diverse outputs based on different sampled masks. Details of baseline methods can be found at Appendix \ref{sec:baseline}.

\vspace{0.02in}
\noindent\textbf{Sampling-based method.} Sampling methods create diverse outputs by sampling next token widely from the vocabulary. We compare against two sampling algorithms for decoding, including truncated sampling~\cite{fan2018hierarchical} and nucleus sampling~\cite{holtzman2020curious}. 
Truncated sampling~\cite{fan2018hierarchical} randomly samples words from top-k probability candidates of the predicted distribution at each decoding step.
Nucleus sampling~\cite{holtzman2020curious} avoids text degeneration by truncating the unreliable tails and sampling from the dynamic nucleus of tokens containing the vast majority of the probability mass.

\subsection{Implementation Details}
\label{sec:imple-details}

All baseline methods were built on the Transformer architecture with 6-layer encoder and decoder, and initialized with pre-trained parameters from BART-base~\cite{lewis2020bart}, which is one of the state-of-the-art pre-trained Transformer models for natural language generation~\cite{gehrmann2021gem}.
In our MoKGE, the Transformer parameters were also initialized by BART-base, in order to make fair comparison with all baseline methods. The R-GCN parameters were random initialized. 

For model training, we used Adam with batch size of 60, learning rate of 3e-5, L2 weight decay of 0.01, learning rate warm up over the first 10,000 steps, and linear decay of learning rate. Our models were trained by one Tesla V100 GPU card with 32GB memory, and implemented on PyTorch with the Huggingface's Transformer~\cite{wolf2020transformers}.
All Transformer-based methods were trained with 30 epochs, taken about 4-5 hours on the ComVE dataset and 7-9 hours on the $\alpha$-NLG dataset.

In addition to our MoKGE implementation, we also provide the baseline implementation code on GitHub \url{https://github.com/DM2-ND/MoKGE}.

\begin{table*}[htb]
\caption{Diversity and quality evaluation on the \textbf{ComVE} (upper part) and \textbf{$\alpha$-NLG} (lower part) datasets. Each model is required to generate three outputs. All experiments are run three times with different random seeds, and the average results on the test set is calculated as the final performance, with standard deviations as
subscripts.}
\vspace{-0.07in}
\begin{subtable}[t]{0.96\textwidth}
\centering
\vspace{-0.04in}
\setlength{\tabcolsep}{2.43mm}{\scalebox{0.85}{\begin{tabular}{c|l||cc|cc|cc|cc}
\toprule
{\multirow{2}*{Methods}} & {\multirow{2}*{\makecell[c]{Model \\ Variant}}}   & \multicolumn{2}{c|}{Concept diversity} & \multicolumn{2}{c|}{Pairwise diversity} & \multicolumn{2}{c|}{Corpus diversity} & \multicolumn{2}{c}{Quality}  \\
\cmidrule{3-10}
& & \#Uni.C($\Uparrow$) & Jaccard ($\Downarrow$) & SB-3 ($\Downarrow$) & SB-4 ($\Downarrow$) & D-2($\Uparrow$) & E-4($\Uparrow$) & B-4 ($\Uparrow$) & R-L ($\Uparrow$) \\
\midrule
{\multirow{3}*{CVAE}} & z\ $=$\ 16 & 4.56$_{\text{0.1}}$ & 64.74$_{\text{0.3}}$ & 66.66$_{\text{0.4}}$ & 62.83$_{\text{0.5}}$ & 33.75$_{\text{0.5}}$ & 9.13$_{\text{0.1}}$ & 16.67$_{\text{0.3}}$ & 41.52$_{\text{0.3}}$  \\
& z\ $=$\ 32 & 5.03$_{\text{0.3}}$ & 47.27$_{\text{0.8}}$ & 59.20$_{\text{1.3}}$ & 54.30$_{\text{1.5}}$ & 32.86$_{\text{1.1}}$ & 9.07$_{\text{0.5}}$ & 17.04$_{\text{0.2}}$ & 42.17$_{\text{0.5}}$  \\
& z\ $=$\ 64 & 4.67$_{\text{0.0}}$ & 54.69$_{\text{0.8}}$ & 55.02$_{\text{0.8}}$ & 49.58$_{\text{1.0}}$ & 32.55$_{\text{0.5}}$ & 9.07$_{\text{0.2}}$ & 15.54$_{\text{0.4}}$ & 41.03$_{\text{0.3}}$ \\
\midrule
{\multirow{3}*{\makecell[c]{Truncated \\ sampling}}} & k\ $=$\ 5 & 4.37$_{\text{0.0}}$ & 71.38$_{\text{0.7}}$ & 74.20$_{\text{0.2}}$ & 71.38$_{\text{0.2}}$ & 31.32$_{\text{0.4}}$ & 9.18$_{\text{0.1}}$ & 16.44$_{\text{0.2}}$ & 40.99$_{\text{0.2}}$  \\
& k\ $=$\ 20 & 4.60$_{\text{0.0}}$ & 63.42$_{\text{1.2}}$ & 64.47$_{\text{2.1}}$ & 60.33$_{\text{2.4}}$ & 33.69$_{\text{0.6}}$ & 9.26$_{\text{0.1}}$ & 17.70$_{\text{0.2}}$ & 42.58$_{\text{0.5}}$  \\
& k\ $=$\ 50 & 4.68$_{\text{0.1}}$ & 60.98$_{\text{1.8}}$ & 61.39$_{\text{2.4}}$ & 56.93$_{\text{2.8}}$ & 34.80$_{\text{0.3}}$ & 9.29$_{\text{0.1}}$ & 17.48$_{\text{0.4}}$ & 42.44$_{\text{0.5}}$  \\
\midrule
{\multirow{3}*{\makecell[c]{Nucleus \\ sampling}}} & p\ $=$\ .5 & 4.19$_{\text{0.1}}$ & 72.78$_{\text{1.0}}$ &77.66$_{\text{0.8}}$ & 75.14$_{\text{0.9}}$ & 28.36$_{\text{0.6}}$ & 9.05$_{\text{0.3}}$ & 16.09$_{\text{0.6}}$ & 40.95$_{\text{0.5}}$ \\
& p\ $=$\ .75 & 4.41$_{\text{0.1}}$ & 67.01$_{\text{1.7}}$ & 71.41$_{\text{2.5}}$ & 68.22$_{\text{2.9}}$ & 31.21$_{\text{0.3}}$ & 9.16$_{\text{0.1}}$ & 17.07$_{\text{0.5}}$ & 41.88$_{\text{0.7}}$  \\
& p\ $=$\ .95 & 4.70$_{\text{0.1}}$ & 61.92$_{\text{2.6}}$ & 63.43$_{\text{3.4}}$ & 59.23$_{\text{3.8}}$ & 34.17$_{\text{0.3}}$ & 9.27$_{\text{0.2}}$ & 17.68$_{\text{0.4}}$ & 42.60$_{\text{0.8}}$  \\
\midrule
{\multirow{2}*{MoE}} & embed & 5.41$_{\text{0.0}}$ & \underline{47.55}$_{\text{0.5}}$ & 33.64$_{\text{0.2}}$ & \underline{28.21}$_{\text{0.1}}$ & 46.57$_{\text{0.2}}$ & 9.61$_{\text{0.1}}$ & 18.66$_{\text{0.5}}$ & \underline{43.72}$_{\text{0.2}}$  \\
& prompt & \underline{5.45}$_{\text{0.2}}$ & 47.54$_{\text{0.4}}$ &
\underline{33.42}$_{\text{0.3}}$ & 28.40$_{\text{0.3}}$ &
46.93$_{\text{0.2}}$ & 9.60$_{\text{0.2}}$ & 18.91$_{\text{0.4}}$ & 43.71$_{\text{0.5}}$  \\ 
\midrule
{\multirow{2}*{\makecell[c]{MoKGE \\ (ours)}}} & embed & 5.35$_{\text{0.2}}$ & 48.18$_{\text{0.5}}$ & 35.36$_{\text{1.1}}$ & 29.71$_{\text{1.2}}$ & \underline{47.51}$_{\text{0.4}}$ & \underline{9.63}$_{\text{0.1}}$ & \textbf{19.13}$_{\text{0.1}}$ & 43.70$_{\text{0.1}}$  \\
& prompt & \textbf{5.48}$_{\text{0.2}}$ & \textbf{44.37}$_{\text{0.4}}$ & \textbf{30.93}$_{\text{0.9}}$ & \textbf{25.30}$_{\text{1.1}}$ & \textbf{48.44}$_{\text{0.2}}$ & \textbf{9.67}$_{\text{0.2}}$ & \underline{19.01}$_{\text{0.1}}$ & \textbf{43.83}$_{\text{0.3}}$  \\
\midrule
Human &  & 6.27$_{\text{0.0}}$ & 26.49$_{\text{0.0}}$ & 12.36$_{\text{0.0}}$ & 8.01$_{\text{0.0}}$ & 63.02$_{\text{0.0}}$ & 9.55$_{\text{0.0}}$ & 100.0$_{\text{0.0}}$ & 100.0$_{\text{0.0}}$  \\
\bottomrule
\end{tabular}}}
\end{subtable}
\begin{subtable}[t]{0.96\textwidth}
\vspace{0.1in}
% \caption{Model performance on \textbf{$\alpha$-NLG} dataset.}
\centering
\vspace{-0.04in}
\setlength{\tabcolsep}{2.43mm}{\scalebox{0.84}{\begin{tabular}{c|l||cc|cc|cc|cc}
\toprule
% {\multirow{2}*{Methods}} & {\multirow{2}*{\makecell[c]{Model \\ Variant}}}   & \multicolumn{2}{c|}{Concept diversity} & \multicolumn{2}{c|}{Pairwise diversity} & \multicolumn{2}{c|}{Corpus diversity} & \multicolumn{2}{c}{Quality}  \\
% \cmidrule{3-10}
& & \#Uni.C($\Uparrow$) & Jaccard ($\Downarrow$) & SB-3 ($\Downarrow$) & SB-4 ($\Downarrow$) & D-2($\Uparrow$) & E-4($\Uparrow$) & B-4 ($\Uparrow$) & R-L ($\Uparrow$) \\
\midrule
{\multirow{3}*{CVAE}} & z\ $=$\ 16 & 4.80$_{\text{0.0}}$ & 56.88$_{\text{0.1}}$ & 67.89$_{\text{0.4}}$ & 64.72$_{\text{0.5}}$ & 26.27$_{\text{0.2}}$ & 10.34$_{\text{0.0}}$ & 13.64$_{\text{0.1}}$ & 37.96$_{\text{0.1}}$ \\
& z\ $=$\ 32 & 5.05$_{\text{0.0}}$ & 50.92$_{\text{0.4}}$ & 62.08$_{\text{0.2}}$ & 58.25$_{\text{0.3}}$ & 26.67$_{\text{0.1}}$ & 10.36$_{\text{0.0}}$ & 13.35$_{\text{0.1}}$ & 37.73$_{\text{0.1}}$  \\
& z\ $=$\ 64 & 5.14$_{\text{0.0}}$ & 47.04$_{\text{0.7}}$ & 57.87$_{\text{0.4}}$ & 53.61$_{\text{0.4}}$ & 24.91$_{\text{0.1}}$ & 10.21$_{\text{0.1}}$ & 11.77$_{\text{0.1}}$ & 36.35$_{\text{0.2}}$  \\
\midrule
{\multirow{3}*{\makecell[c]{Truncated \\ sampling}}} & k$=$\ 5 & 4.86$_{\text{0.1}}$ & 72.78$_{\text{1.1}}$ & 67.09$_{\text{1.0}}$ & 63.82$_{\text{1.1}}$ & 25.47$_{\text{0.3}}$ & 10.44$_{\text{0.1}}$ & 13.33$_{\text{0.2}}$ & 38.07$_{\text{0.2}}$  \\
& k$=$\ 20 & 5.48$_{\text{0.1}}$ & 45.65$_{\text{1.8}}$ & 54.65$_{\text{2.1}}$ & 50.36$_{\text{2.4}}$ & 29.30$_{\text{0.5}}$ & 10.62$_{\text{0.2}}$ & 14.12$_{\text{0.7}}$ & 38.76$_{\text{0.6}}$  \\
& k$=$\ 50 & 5.53$_{\text{0.0}}$ & 45.84$_{\text{0.5}}$ & 52.11$_{\text{3.7}}$ & 47.75$_{\text{4.2}}$ & 30.08$_{\text{0.3}}$ & 10.64$_{\text{0.1}}$ & 14.01$_{\text{0.8}}$ & \textbf{38.98}$_{\text{0.6}}$ \\
\midrule
{\multirow{3}*{\makecell[c]{Nucleus \\ sampling}}} & p$=$\ .5 & 4.19$_{\text{0.1}}$ & 62.54$_{\text{1.8}}$ & 73.34$_{\text{0.3}}$ & 71.01$_{\text{0.3}}$ & 25.49$_{\text{0.0}}$ & 10.46$_{\text{0.0}}$ & 11.71$_{\text{0.1}}$ & 36.53$_{\text{0.2}}$ \\
& p$=$\ .75 & 5.13$_{\text{0.0}}$ & 54.25$_{\text{0.6}}$ & 64.49$_{\text{0.4}}$ & 61.45$_{\text{0.5}}$ & 27.72$_{\text{0.1}}$ & 10.54$_{\text{0.1}}$ & 12.63$_{\text{0.0}}$ & 37.48$_{\text{0.1}}$  \\
& p$=$\ .95 & 5.49$_{\text{0.0}}$ & 46.76$_{\text{0.5}}$ & 56.32$_{\text{0.5}}$ & 52.44$_{\text{0.6}}$ & 29.92$_{\text{0.1}}$ & 10.63$_{\text{0.0}}$ & 13.53$_{\text{0.2}}$ & 38.42$_{\text{0.3}}$  \\
\midrule
{\multirow{2}*{MoE}} & embed & 6.22$_{\text{0.1}}$ & \underline{29.18}$_{\text{0.4}}$ & 29.02$_{\text{1.0}}$ & 24.19$_{\text{1.0}}$ & 36.22$_{\text{0.3}}$ & 10.84$_{\text{0.0}}$ & \textbf{14.31}$_{\text{0.2}}$ & \underline{38.91}$_{\text{0.2}}$ \\
& prompt & 6.05$_{\text{0.1}}$ & 29.34$_{\text{1.2}}$ & \underline{28.05}$_{\text{2.0}}$ & \underline{23.18}$_{\text{1.9}}$ & 36.71$_{\text{0.1}}$ & 10.85$_{\text{0.0}}$ & \underline{14.26}$_{\text{0.3}}$ & 38.78$_{\text{0.4}}$  \\
\midrule
{\multirow{2}*{\makecell[c]{MoKGE \\ (ours)}}} & embed & \underline{6.27}$_{\text{0.2}}$ & 30.46$_{\text{0.8}}$ & 29.17$_{\text{1.5}}$ & 24.04$_{\text{1.6}}$ & \textbf{38.15}$_{\text{0.3}}$ & \textbf{10.90}$_{\text{0.1}}$ & 13.74$_{\text{0.2}}$ & 38.06$_{\text{0.2}}$  \\
& prompt & \textbf{6.35}$_{\text{0.1}}$ & \textbf{28.06}$_{\text{0.6}}$ & \textbf{27.40}$_{\text{2.0}}$ & \textbf{22.43}$_{\text{2.4}}$ & \underline{38.01}$_{\text{0.6}}$ & \underline{10.88}$_{\text{0.2}}$ & 14.17$_{\text{0.2}}$ & 38.82$_{\text{0.7}}$  \\
\midrule
Human & & 6.62$_{\text{0.0}}$ & 12.43$_{\text{0.0}}$ & 10.36$_{\text{0.0}}$ & 6.04$_{\text{0.0}}$ & 53.57$_{\text{0.0}}$ & 10.84$_{\text{0.0}}$ & 100.0$_{\text{0.0}}$ & 100.0$_{\text{0.0}}$  \\
\bottomrule
\multicolumn{10}{l}{* Metrics: SB-3/4: Self-BLEU-3/4 ($\Downarrow$), D-2: Distinct-2 ($\Uparrow$), E-4: Entropy-4 ($\Uparrow$), B-4: BLEU-4 ($\Uparrow$), R-L: ROUGE-L ($\Uparrow$)}\\
\end{tabular}}}
\end{subtable}
\label{tab:baseline-1}
\end{table*}

\begin{table*}[htb]
\begin{center}
\caption{Ablation studies. When not suing MoE (line --w/o MoE), we set beam as three to generate three outputs.}
\vspace{-0.1in}
\scalebox{0.83}{\begin{tabular}{l||ccc|cc|ccc|cc}
\toprule
{\multirow{3}*{\makecell[c]{Methods}}} & \multicolumn{5}{c|}{ComVE (left part: diversity; right part: quality)} & \multicolumn{5}{c}{$\alpha$-NLG (left part: diversity; right part: quality)} \\
\cmidrule{2-11}
& SB-4 ($\Downarrow$) & D-2 ($\Uparrow$) & E-4 ($\Uparrow$) & B-4 ($\Uparrow$) & R-L ($\Uparrow$) & SB-4 ($\Downarrow$) & D-2 ($\Uparrow$) & E-4 ($\Uparrow$) & B-4 ($\Uparrow$) & R-L ($\Uparrow$) \\
\midrule 
MoKGE & \textbf{25.30}$_{\text{1.1}}$ & \textbf{48.44}$_{\text{0.2}}$ & \textbf{9.67}$_{\text{0.2}}$ & \textbf{19.01}$_{\text{0.1}}$ & \textbf{43.83}$_{\text{0.3}}$ & \textbf{22.43}$_{\text{2.4}}$ & \textbf{38.01}$_{\text{0.6}}$ & \textbf{10.88}$_{\text{0.2}}$ & 14.17$_{\text{0.2}}$ & \textbf{38.82}$_{\text{0.7}}$ \\
~$\vdash$ w/o KG & 28.40$_{\text{0.3}}$ & 46.93$_{\text{0.2}}$ & 9.60$_{\text{0.2}}$ & 18.91$_{\text{0.4}}$ & 43.71$_{\text{0.5}}$ & 23.18$_{\text{1.9}}$ & 36.71$_{\text{0.1}}$ & 10.85$_{\text{0.0}}$ & \textbf{14.26}$_{\text{0.3}}$ & 38.78$_{\text{0.4}}$ \\
~$\vdash$ w/o MoE & 74.15$_{\text{0.2}}$ & 31.92$_{\text{0.1}}$ & 9.14$_{\text{0.0}}$ & 15.87$_{\text{0.1}}$ & 40.24$_{\text{0.2}}$ & 77.34$_{\text{0.2}}$ & 19.19$_{\text{0.1}}$ & 10.10$_{\text{0.0}}$ & 12.84$_{\text{0.1}}$ & 37.52$_{\text{0.2}}$   \\
\bottomrule
\end{tabular}}
\label{tab:ablation}
\end{center}
\vspace{-0.1in}
\end{table*}

\begin{table*}[t]
\begin{center}
\caption{Human evaluations by independent scoring based on \textit{diveristy}, \textit{quality}, \textit{flency} and \textit{grammar}. In addition, \\ * indicates $p$-value $<0.05$ under paired $t$-test between MoKGE and baseline methods.}
\vspace{-0.1in}
\setlength{\tabcolsep}{3mm}{\scalebox{0.9}{\begin{tabular}{l||lcc|lcc}
\toprule
{\multirow{2}*{\makecell[c]{Methods}}} & \multicolumn{3}{c|}{ComVE} & \multicolumn{3}{c}{$\alpha$-NLG} \\
\cmidrule{2-7}
& Diversity & Quality & Flu. \& Gra. & Diversity & Quality & Flu. \& Gra.  \\
% \midrule
% \rowcolor{gray!12}\multicolumn{4}{c}{ComVE task} \\
\midrule
% {\multirow{1}*{{Beam search}}} & 1.61$\pm$0.74 & 2.28$\pm$0.90 & 3.48$\pm$0.84 \\
{\multirow{1}*{{Truncated samp.}}} & 2.15$\pm$0.76 & 2.22$\pm$1.01 & 3.47$\pm$0.75 & 2.31$\pm$0.76 & 2.63$\pm$0.77 & 3.89$\pm$0.36 \\
{\multirow{1}*{Nucleus samp.}} & 2.03$\pm$0.73 & \textbf{2.29}$\pm$1.03 & \textbf{3.52}$\pm$0.70 & 2.39$\pm$0.73 & \textbf{2.67}$\pm$0.72 & \textbf{3.91}$\pm$0.28 \\
{\multirow{1}*{MoKGE (ours)}} & \textbf{2.63}$\pm$0.51* & 2.10$\pm$0.99 & 3.46$\pm$0.81 & \textbf{2.66}$\pm$0.51* & 2.57$\pm$0.71 & 3.87$\pm$0.34 \\
{\multirow{1}*{{Human Ref.}}} & 2.60$\pm$0.59 & 3.00 & 4.00 & 2.71$\pm$0.57 & 3.00 & 4.00 \\
\bottomrule
\end{tabular}}}
\vspace{-0.05in}
\label{tab:human-eval-2}
\end{center}
\begin{center}
\caption{Human evaluations by pairwise comparison: MoKGE v.s. two baseline methods based on \textit{diversity}.}
\vspace{0.02in}
\scalebox{0.91}{\begin{tabular}{l||ccr|ccc}
\toprule
{\multirow{2}*{\makecell[c]{Against methods}}} & \multicolumn{3}{c|}{ComVE} & \multicolumn{3}{c}{$\alpha$-NLG} \\
\cmidrule{2-7}
& Win (\%) & Tie (\%) & Lose (\%) & Win (\%) & Tie (\%) & Lose (\%) \\
\midrule
{\multirow{1}*{v.s. Truncated samp.}} & \textbf{47.85}$\pm$5.94 & 37.09$\pm$4.56 & 15.06$\pm$3.31 & \textbf{45.35}$\pm$5.06 & 43.19$\pm$2.78 & 11.46$\pm$2.31 \\
{\multirow{1}*{v.s. Nucleus samp.}} & \textbf{54.30}$\pm$4.62 & 36.02$\pm$2.74 & 9.68$\pm$3.48 & 41.53$\pm$1.55 & \textbf{46.99}$\pm$2.04 & 11.48$\pm$2.36 \\
\bottomrule
\end{tabular}}
\vspace{-0.2in}
\label{tab:human-eval}
\end{center}
\end{table*}

% \begin{table*}[t]
% \begin{center}
% \caption{Human evaluations by pairwise comparison: MoKGE v.s. two baseline methods based on \textit{diversity}. }
% \vspace{-0.1in}
% \scalebox{0.9}{\begin{tabular}{l||ccr|ccc}
% \toprule
% {\multirow{2}*{\makecell[c]{Against methods}}} & \multicolumn{3}{c|}{ComVE} & \multicolumn{3}{c}{$\alpha$-NLG} \\
% \cmidrule{2-7}
% & Win (\%) & Tie (\%) & Lose (\%) & Win (\%) & Tie (\%) & Lose (\%) \\
% \midrule
% {\multirow{1}*{v.s. Truncated samp.}} & \textbf{47.85}$\pm$5.94 & 37.09$\pm$4.56 & 15.06$\pm$3.31 & \textbf{45.35}$\pm$5.06 & 43.19$\pm$2.78 & 11.46$\pm$2.31 \\
% {\multirow{1}*{v.s. Nucleus samp.}} & \textbf{54.30}$\pm$4.62 & 36.02$\pm$2.74 & 9.68$\pm$3.48 & 41.53$\pm$1.55 & \textbf{46.99}$\pm$2.04 & 11.48$\pm$2.36 \\
% \bottomrule
% \end{tabular}}
% \vspace{-0.2in}
% \label{tab:human-eval}
% \end{center}
% \end{table*}

\subsection{Automatic Evaluation}

We evaluated the performance of different generation models from two aspects: \textit{quality} (or say \textit{accuracy}) and \textit{diversity}. \textit{Quality} tests the appropriateness of the generated response with respect to the context, and \textit{diversity} tests the lexical and semantic diversity of the appropriate sequences generated by the model. These evaluation metrics have been widely used in existing work~\cite{ott2018analyzing,vijayakumar2018diverse,zhu2018texygen,cho2019mixture,yu2021sentence}.

% \vspace{-0.05in}
% \subsubsection{Quality metrics}
% Computing word-overlap metrics against the same references may penalize other reasonable output sequence~\cite{deriu2020survey,gupta2019investigating}. 
% % We use multi-reference BLEU (mBLEU) (aka. coverage, recall~\cite{zhao2017learning,gupta2018deep}).
% Given multiple reference responses $R =
% \{r_1, r_2, \cdots, r_n\}$ and multiple generated outputs $Y = \{y_1, y_2, ..., y_m\}$. The oracle BLEU (short as O-BLEU) is calculated as: 
% \begin{equation}
%     \text{O-BLEU} = \frac{\sum_{j=1}^{m} \max_{i \in [1, n]}{\text{BLEU} (y_i, r_j)}}{m},
% \end{equation}
% For each generated sequence, we consider the highest-scoring output, then average these scores across the reference responses. A system that generates outputs covering a large portion of the reference responses thus receives a higher recall score.

\vspace{0.03in}
\noindent\textbf{Quality metrics ($\Uparrow$).} The quality is measured by standard N-gram based metrics, including the BLEU score~\cite{papineni2002bleu} and the ROUGE score~\cite{lin2004rouge}. 
This measures the highest accuracy comparing the best hypothesis among the top-$K$ with the target~\cite{vijayakumar2018diverse}. 
Concretely, we generate hypotheses $\{{\hat{Y}^{(1)}, \cdots \hat{Y}^{(K)}}\}$ from each source $X$ and keep the hypothesis $\hat{Y}^{\text{best}}$ that achieves the best sentence-level metric with the target $Y$. Then we calculate a corpus-level metric with the greedily-selected hypotheses $\{{Y^{(i),\text{best}}\}}^N_{i=1}$ and references $\{{Y^{(i)}}\}^N_{i=1}$.

\vspace{0.03in}
The diversity of evaluated by three aspects: concept, pairwise and corpus diversity.

\vspace{0.02in}
\noindent\textbf{Concept diversity.}
The number of unique concepts (short as Uni.C) measures how many unique concepts on the commonsense KG are covered in the generated outputs. A higher value indicates the higher concept diversity. Besides, we also measure the pairwise concept diversity by using Jaccard similarity. It is defined as the size of the intersection divided by the size of the union of two sets. 
% Given multiple outputs, we calculate the mean value of Jaccard of all pairwise combinations (in total $C_N^2$). 
Lower value indicates the higher concept diversity.
% \vspace{-0.05in}
% \begin{equation}
%     \text{Avg-Jaccard} = \frac{1}{C_N^2} \sum_{i=1}^{N-1} \sum_{j=i+1}^{N} \text{Jaccard}(O_i, O_j), \nonumber
% \end{equation}
% where $O_i$ is the $i$-th outputs. Lower value is better.

\vspace{0.02in}
\noindent\textbf{Pairwise diversity ($\Downarrow$).} Referred as ``self-'' (e.g., self-BLEU)~\cite{zhu2018texygen}, it measures the within-distribution similarity. This metric computes the average of sentence-level metrics between all pairwise combinations of hypotheses $\{{Y^{(1)}, \cdots, Y^{(K)}}\}$ generated from each source sequence $X$. Lower pairwise metric indicates high diversity between generated hypotheses.

\vspace{0.02in}
\noindent\textbf{Corpus diversity ($\Uparrow$).}
Distinct-$k$~\cite{li2016diversity} measures the total number of unique $k$-grams normalized by the total number of generated $k$-gram tokens to avoid favoring long sentences. Entropy-$k$~\cite{zhang2018generating} reflects how evenly the empirical $k$-gram distribution is for a given sentence when word frequency is considered.

% N-gram based metrics are sensitive to the lexical diversity, rather than the content-level diversity~\cite{tevet2020evaluating}. Therefore, in addition of using Self-BLEU, we also adopt recent embedding based metric BERT-score~\cite{zhang2020bertscore} to evaluate the content-level diversity, which is referred as Self-BERT in our paper.

\begin{figure*}[t]
    \centering
    {\includegraphics[width=1.0\textwidth]{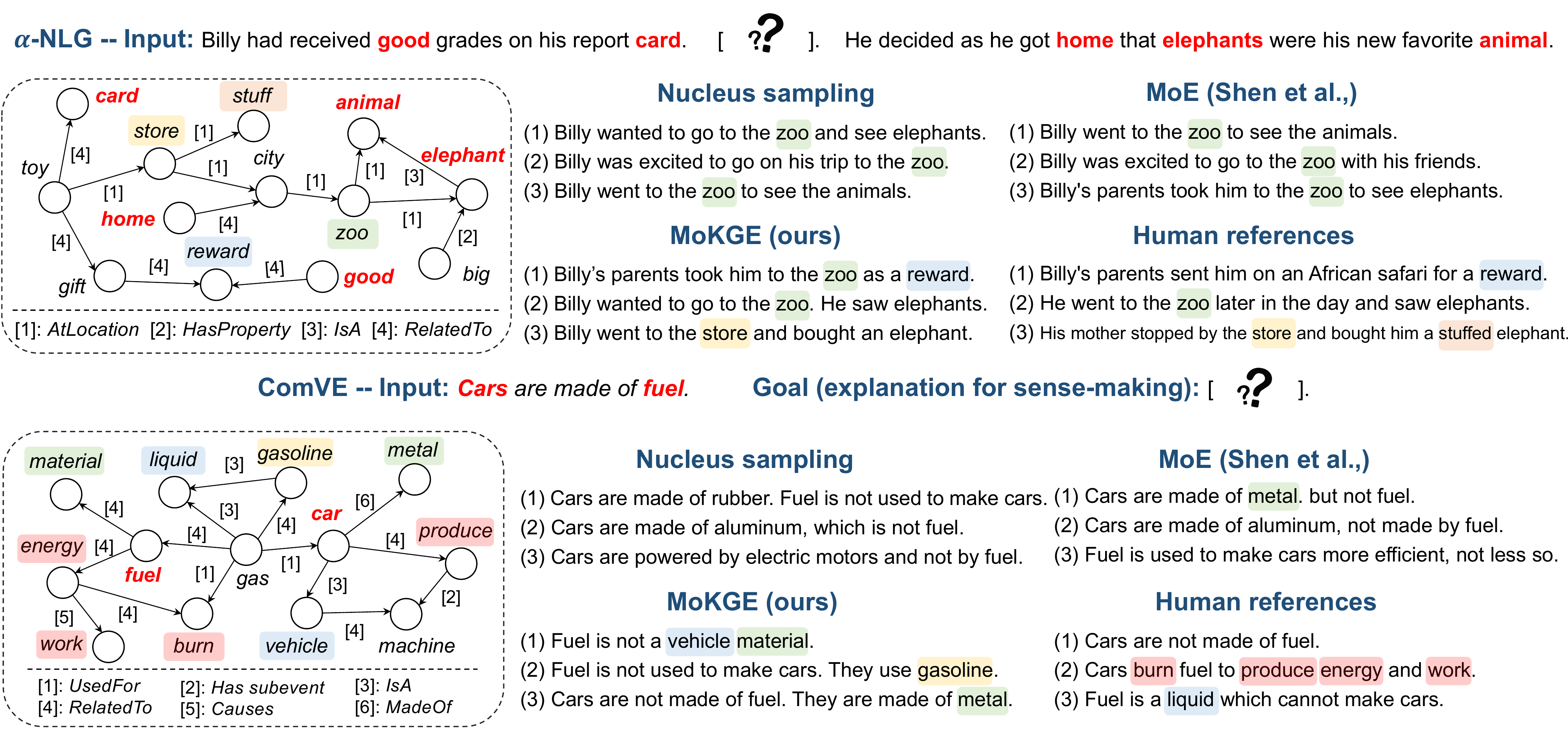}}
    \vspace{-0.25in}
    \caption{Case studies. MoKGE can produce diverse knowledge reasoning on commonsense KG, select different relevant concepts (in shades of different colors), then generate diverse outputs. The outputs diversity of MoKGE is significantly better than that of beam search and nucleus sampling, and close to human performance.}
    \label{fig:case}
\end{figure*}

\subsubsection{Experimental results}

\noindent\textbf{Comparison with baseline methods.}
We evaluated our proposed MoKGE and baseline methods based on both \textit{quality} and \textit{diversity}.
As shown in Table \ref{tab:baseline-1}, MoE-based methods achieved the best performance among all baseline methods. 
MoKGE can further boost diversity by at least 1.57\% and 1.83\% on Self-BLEU-3 and Self-BLEU-4, compared with the vanilla MoE methods. At the same time, MoKGE achieved on par performance with other baseline methods based on the quality evaluation. 
Specifically, on the ComVE dataset, MoKGE achieved the best performance on BLEU-4 and ROUGE-L, and on the $\alpha$-NLG dataset, the performance gap between MoKGE and the best baseline method was always \textit{less than 0.5\%} on BLEU-4.

\vspace{0.02in}
\noindent\textbf{Ablation study.}
We conducted an ablation study to analyze the two major components in the MoKGE. The experimental results are shown in Table \ref{tab:ablation}.
First, we note that when not using MoE (line --w/o MoE), we used the most basic decoding strategy -- beam search -- to generate multiple outputs. 
We observed that the outputs generated by beam search differed only on punctuation and minor morphological variations, and typically only the last few words were different from others. 
Besides, integrating commonsense knowledge graph into the MoE-based generation model brought both quality and diversity improvement on the ComVE, but might sacrifice a little quality (less than 0.5\% on BLEU-4) on the $\alpha$-NLG dataset.
Overall, our MoKGE benefited from KG and MoE modules, and achieved great performance on both diversity and quality.

% \vspace{0.02in}
% \noindent\textbf{Trade-off analysis}
% % \red{TODO maybe one more figure is needed!} 
% We analyzed the trade-off between generation quality and diversity in different training epochs on the validation set.
% As shown in Figure \ref{fig:trade-off}, we observed the performance trend of the quality evaluation and diversity evaluation are out of sync. As training proceeds, the quality of the generated outputs first increase and then slightly decreases. However, the diversity performance continues to improve until it reaches a certain value.
% Therefore, in order to guarantee the model to produce multiple outputs with relatively good quality and diversity, we choose epochs with top-5 quality performance on the validation set and pick the epoch with the best diversity for testing. 

\subsection{Human Evaluation}

Automatic diversity evaluation (e.g., Self-BLEU, Distinct-$k$) cannot reflect the \textit{content-level diversity}. 
Therefore, we conducted extensive human evaluations to assess both the quality and diversity of outputs generated from different models. 

The human evaluation was divided into two parts: \textit{independent scoring} and \textit{pairwise comparisons}. All evaluations were conducted on Amazon Mechanical Turk (AMT), and each evaluation form was answered by at least three AMT workers.

\vspace{0.05in}
\noindent\textbf{Independent scoring.} In this part, human annotators were asked to evaluate the generated outputs from a single model. 
We first presented top-3 generated outputs from a certain model to human annotators. The annotators would first evaluate the \textit{diversity} by answering ``How many different meanings do three outputs express?'' Then we presented human-written outputs to the annotators. The annotator would evaluate the quality by comparing machine generated outputs and human-written outputs, and answering ``How many machine generated outputs are correct?'' The diversity and quality scores are normalized to the range from 0 to 3.
Besides, the annotators need to give a fluency and grammar score from 1 to 4 for each generated output.

\vspace{0.05in}
\noindent\textbf{Pairwise comparisons.} In this part, the annotators were given two sets of top-3 generated explanations from two different methods each time and instructed to pick the more diverse set. The choices are ``win,'' ``lose,'' or ``tie.''

As shown in Table~\ref{tab:human-eval-2}-\ref{tab:human-eval}, our MoKGE can significantly outperform the state-of-the-art sampling-based methods in \textit{diversity} evaluation ($p$-value $ < 0.05$ under paired $t$-test), even slightly better than human performance on the ComVE task. At the same time, we can observe MoKGE is able to obtain on par performance with other methods based on \textit{quality} evaluation.
The $p$-value is not smaller than $0.05$ (i.e., not significant difference) under paired $t$-test between MoKGE and baseline methods based on the quality evaluation.

\subsection{Case Study} 

Figure \ref{fig:case} demonstrates human-written explanations and generated explanations from different diversity-promoting methods, including nucleus sampling, mixture of experts (MoE) and our MoKGE.
Overall, we observed that the nucleus sampling and MoE methods typically expressed very similar meanings, e.g., ``go to the zoo and see elephants'' and ``took him to the zoo and see elephants'' in the $\alpha$-NLG case. 
On the contrary, MoKGE can generate semantically richer and more diverse contents than the other two methods by incorporating more commonsense concepts on the knowledge graph.
% We notice that the explanations generated by beam search differ only on punctuation and minor morphological variations, and typically only the last few words are different from others. 
% MoE achieves better diversity than beam search, but the explanations still convey very similar meanings, e.g., ``go to the zoo and see elephants'' and ``took him to the zoo and see elephants'' in the $\alpha$-NLG case. 
% In comparison, MoKGE can generate semantically richer and more diverse contents.

%% file: 5future.tex
\vspace{0.02in}
\noindent\textbf{Improving content diversity in NLG.}
Most of the existing diversity-promoting work has focused on improving syntactic and lexical diversity, such as different language style in machine translation~\cite{shen2019mixture} and word variability in paraphrase generation~\cite{gupta2018deep}. 
Nevertheless, methods for improving content diversity in NLG systems have been rarely studied in the existing literature. We believe that generating diverse content is one of the most promising aspects of machine intelligence, which can be applied to a wide range of real-world applications, not only limited to commonsense reasoning.

Besides, leveraging knowledge graph is not the only way to promote content diversity as it is a highly knowledge-intensive task. Many existing knowledge-enhanced methods~\cite{yu2020survey} can be used to acquire different external knowledge for producing diverse outputs, e.g., taking different retrieved documents as conditions for generator.

\vspace{0.05in}
\noindent\textbf{Designing neural diversity metrics.}
In spite of growing interest in NLG models that produce diverse outputs, there is currently no principled neural method for evaluating the diversity of an NLG system. 
As described in \citet{tevet2021evaluating}, existing automatic diversity metrics (e.g. Self-BLEU) perform worse than humans on the task of estimating content diversity, indicating a low correlation between metrics and human judgments.

Therefore, neural-based diversity metrics are highly demanded. Intuitively, the metrics should include computational comparisons of multiple references and hypotheses by projecting them into the same semantic space, unlike metrics for evaluating the generation quality, e.g., BERTScore~\cite{zhang2020bertscore} and BLEURT~\cite{sellam2020bleurt}, which only measures the correlation between a pair of reference and hypothesis.

%% file: 5conclusions.tex
In this paper, we proposed a novel method that diversified the generative reasoning by a mixture of expert strategy on commonsense knowledge graph.
To the best of our knowledge, this is the first work to boost diversity in NLG by diversifying knowledge reasoning on commonsense knowledge graph.
Experiments on two generative commonsense reasoning benchmarks demonstrated that MoKGE outperformed state-of-the-art methods on diversity, while achieving on par performance on quality.